# An enhanced Teaching-Learning-Based Optimization (TLBO) with Grey Wolf Optimizer (GWO) for text feature selection and clustering

Mahsa Azarshab[1], Mohammad Fathian[1], Babak Amiri[1]

**Abstract**

Text document clustering can play a vital role in organizing and handling the ever-increasing number of text documents. Uninformative and redundant features included in large text documents reduce the effectiveness of the clustering algorithm. Feature selection (FS) is a well-known technique for removing these features. Since FS can be formulated as an optimization problem, various meta-heuristic algorithms have been employed to solve it. Teaching-Learning-Based Optimization (TLBO) is a novel meta-heuristic algorithm that benefits from the low number of parameters and fast convergence. A hybrid method can simultaneously benefit from the advantages of TLBO and tackle the possible entrapment in the local optimum. By proposing a hybrid of TLBO, Grey Wolf Optimizer (GWO), and Genetic Algorithm (GA) operators, this paper suggests a filter-based FS algorithm (TLBO-GWO). Six benchmark datasets are selected, and TLBO-GWO is compared with three recently proposed FS algorithms with similar approaches, the main TLBO and GWO. The comparison is conducted based on clustering evaluation measures, convergence behavior, and dimension reduction, and is validated using statistical tests. The results reveal that TLBO-GWO can significantly enhance the effectiveness of the text clustering technique (K-means).

**Keywords** Text clustering. Feature selection. Teaching-learning-based optimization. Grey wolf optimizer. Genetic operators

[1] School of Industrial Engineering, Iran University of Science and Technology, Narmak, Tehran, Iran Postal code: 16846-13114, Fax: +982173225098.
Email addresses: mahsa_azarshab@alumni.iust.ac.ir (M. Azarshab), fathian@iust.ac.ir (M. Fathian), babakamiri@iust.ac.ir (B. Amiri).

# 1. Introduction

Over the past few decades, because of the growth of websites and modern applications, as well as their users, unstructured text data has been produced on a massive scale. Text clustering is an effective technique in text mining that can organize innumerable text documents and partition them into a predetermined number of coherent clusters containing the most similar documents1]. In the text mining tasks, texts are represented as weighted vectors of terms. Each term is considered a feature and a dimension in this representation. Values of weighted feature vectors are usually computed using the Term Frequency-Inverse Document Frequency (TF-IDF). Therefore, each document is characterized in a multi-dimensional vector, and inevitably, text datasets should be shown as high-dimensional arrays containing both informative and uninformative features. The large size of text documents, as well as the existence of uninformative features, weaken the performance of the clustering algorithm.

Feature selection (FS) is a well-known technique that is able to eliminate uninformative, noisy, redundant, and irrelevant features and find a subset of informative features. In literature, various classifications have been introduced regarding FS methods. A well-known categorization of these methods is based on the mechanism of assessing the features; accordingly, the two main FS approaches are filter and wrapper models.

Filters use general data characteristics to select the optimal subset of features and do not require a training algorithm in the FS process. These methods use different criteria to evaluate the features and assign a score to each subset of features. Document frequency [2], term frequency-inverse document frequency [3], mutual information [4], information gain [5], term variance [6], the mean absolute difference [6], and the Gini index [7] are some of the evaluation criteria. Wrapper models need a training algorithm to examine the selected subset of features. Although wrappers can choose the features more effectively and lead to more accurate clustering results, they suffer from high computational costs. It is noteworthy that filter methods are more computationally effective than wrappers. Due to their simplicity and efficiency, filter models have been used in many FS methods in the text clustering application. It should be noted that it is essential to employ an effectual FS approach to extract the smallest subset of the most informative features.

FS can be formulated as an optimization problem and solved by meta-heuristic algorithms. This paper introduces a new filter-based FS algorithm by presenting a novel Teaching-Learning-Based Optimization (TLBO-GWO). TLBO-GWO is designed based on Teaching-Learning-Based Optimization (TLBO), Grey Wolf Optimizer (GWO), and Genetic Algorithm (GA) operators. The main objective of this paper is to propose a new FS algorithm to enhance the results of text clustering. After selecting the subset of features, the K-means clustering method is used to cluster the updated text dataset. Five comparative algorithms are also implemented to be compared with the proposed one. The results are evaluated based on four clustering measures and validated using statistical tests. Furthermore, algorithms are compared based on convergence behavior and dimension reduction capability.

This paper is organized as follows: some recent works that employed meta-heuristics to solve the FS problem are described in Section 2. The details of the proposed methodology are explained in Section 3. Datasets, evaluation measures, and setups are introduced in Section 4. Finally, results and conclusions are provided in Sections 5 and 6, respectively.

## 2. Literature review

Various works have been carried out to solve the FS problem so as to enhance the results of data mining tasks. Since FS can be defined as an optimization problem, meta-heuristic algorithms have been widely applied to find an optimal solution for this problem.

As mentioned earlier, filters and wrappers are two major categories of FS methods. Although wrappers have a high computational cost, they can produce better clustering results. Mafarja et al. proposed a wrapper-based FS method using Particle Swarm Optimization (PSO) [9,10]. This work examined the effect of different updating strategies for inertia weight to achieve a proper balance between exploration and exploitation. A wrapper-based FS approach was introduced by Allam et al. in which a binary Teaching-Learning-Based Optimization (TLBO) was suggested to find the optimal subset of features [11]. This approach benefited from the small number of parameters and eliminated parameter-tuning challenges. Mafarja et al. used a Whale Optimization Algorithm (WOA) [12] with two different updating approaches to select the optimal subset of features [13]. The first approach used tournament and roulette wheel selection mechanisms to improve the search process. The second approach employed genetic operators to enhance the exploitation power. Another wrapper-based unsupervised FS method was proposed based on a modified version of the Binary Bat Algorithm (BBA) [14]. Then K-means was used to evaluate the performance of the presented method. A multi-objective wrapper-based framework was suggested by Prakash et al. based on Gravitational Search Algorithm (GSA) [15] and K-means to solve FS problems [16]. Abualigah et al. proposed a new hybrid FS method using the Sine Cosine Algorithm (SCA) [17] and Genetic Algorithm (GA) in order to achieve the minimum number of features and maximum classification accuracy [18]. The proposed approach used GA to design an enhanced version of SCA with a proper balance between exploration and exploitation. In [19], a multi-objective FS algorithm was suggested using the Forest Optimization Algorithm (FOA) [20]. Two versions of the introduced algorithm were designed based on continuous and binary representations. Samieiyan et al. [21] presented a novel FS method by introducing an enhanced version of the Crow Search Algorithm (CSA) [22]. The proposed approach kept a proper balance between exploitation and exploration using dynamic awareness probability and prevented entrapment in the local optimum. Another wrapper-based FS method was suggested by Khosravi et al. [23] based on a novel Group Teaching Optimization Algorithm (GTOA) [24]. They improved the main GTOA with local search and chaos mapping. In addition, several operators were applied to enhance the main algorithm's exploration, exploitation, and convergence ability.

Even though filter-based methods might not lead to as accurate clustering results as wrappers, they have lower computational costs and are more appropriate for high-dimensional data [25]. Shamsinejad et al. introduced a new GA-based FS method in which groups of terms were processed instead of every single term [26]. Moreover, they proposed a modified version of term variance to evaluate the features. Sung-Sam Hong et al. designed a new genetic algorithm to select an optimized subset of features [27]. In this research, TF-IDF was employed to assess the importance of features. Then they used spam mail documents and a clustering approach to evaluate the proposed method. Abualigah et al. introduced a new FS method (FSPSOTC) using Particle Swarm Optimization (PSO) [28]. TF-IDF was used as a weighting scheme, and Mean Absolute Difference (MAD) was employed to evaluate each solution. Experiments were conducted using six benchmark datasets afterward. Abualigah et al. [6] used genetic operators to enhance the performance of the PSO algorithm and proposed a new FS method (H-FSPSOTC). A hybrid FS algorithm was suggested by Purushothaman et al. [3] based on Grey Wolf Optimizer (GWO) [29] and Grasshopper Optimization Algorithm (GOA) [30]. GWO was able to select local optimum features, and GOA was considered to choose the best global optimums. After conducting the feature selection process, fuzzy C-Means clustering was used to evaluate the quality of the selected subset of features. Abualigah et al. enhanced the Krill Herd (KH) algorithm [31] by hybridizing with the swap mutation strategy and incorporating it within the parallel membrane framework [32]. A new binary PSO was proposed by Kushwaha et al. to solve the FS problem [33]. They introduced a new updating strategy in which the neighbor best position was used instead of the global best, and eventually, the K-means clustering algorithm evaluated the obtained features. Abualigah et al. suggested a new FS technique based on a Harmony Search (HS) [34] to improve text clustering effectiveness [35]. Thiyagarajan et al. proposed a modification of the Artificial Fish Swarm algorithm (AFSA) [36] and took the advantage of crossover operator in order to design a multi-objective feature selection algorithm and improve the text categorization [37].

By studying the literature, it can be inferred that meta-heuristic algorithms can play a significant role in solving the FS problem and enhancing text clustering results. These algorithms dispense with exploring the whole search area; consequently, an informative subset of features could be achieved in a significantly shorter time. These results encourage examining the effectiveness of other novel meta-heuristic algorithms to produce a proper solution for the FS problem.

TLBO is a recently developed meta-heuristic algorithm proposed by Rao et al. [38]. It is inspired by the classroom's teaching and learning process and has been recently employed to solve various optimization problems. Since TLBO has a low number of parameters and does not deal with parameter-tuning issues, it has attracted the attention of researchers. Despite all its advantages, some research reported that TLBO might suffer premature convergence in some cases [38].

Over the past few years, hybrid meta-heuristic algorithms have attracted considerable attention among researchers. Hybrid methods can benefit from the advantages of multiple algorithms and reach a substantial balance between exploration and exploitation. Mirjalili et al. introduced a new meta-heuristic algorithm called Grey Wolf Optimizer (GWO) [29], inspired by grey wolves' leadership hierarchy and hunting mechanism [29]. This algorithm has been applied to a wide range of optimization problems and has shown a remarkable convergence ability toward the optimum solution. Therefore, in this study, a hybrid algorithm is proposed using TLBO, GWO, and genetic operators to find a more informative subset of features and improve the performance of text clustering.

## 3. Proposed methodology

A novel method is proposed based on TLBO, GWO, and genetic operators to find a subset of features leading to more accurate results of text clustering. At the first level of the proposed methodology, text pre-processing should be conducted to represent a collection of text documents in a standard numerical format. A hybrid algorithm of TLBO and GWO (TLBO-GWO), boosted with genetic operators, is applied at the second level to select the most informative subset of features. It is worth mentioning that the proposed algorithm is implemented at the level of each document. Therefore, TLBO-GWO is executed document by document until the maximum number of iterations. Afterwards, the selected features of all documents are combined to build a global subset of features. Eventually, the collection of text documents is updated based on the selected features and then is partitioned into optimal clusters using the K-means algorithm. The details of the proposed approach are described in the following subsections.

### 3.1. Text pre-processing

Pre-processing steps are necessary actions that should be taken to demonstrate text documents in a standard numerical format. The following subsections explain some significant pre-processing steps.

#### 3.1.1. Tokenization

Each text document contains numerous words and spaces. Tokenization is the task of splitting each sentence into its constituent words, in which each word is considered a token.

#### 3.1.2. Stop words removal

"Stop word" refers to those common words like a, an, the, if, this, that, and other highly frequented, uninformative, and unfunctional words in the text collection. The inclusion of these words in a dataset can weaken the clustering algorithm and reduce its accuracy.

#### 3.1.3. Stemming

Stemming is the stage of extracting words' stem (root) by removing their suffixes and prefixes. It transforms inflectionally relevant forms of a word to the same root. For example, the common root of "connect," "connection," "connected," and "connects" is "connect."

### 3.1.4. Term weighting

In the final step, converting the textual format to numerical and vector representation is necessary. This representation is called the "Vector Space Model" (VSM) and represents each document as a vector of weights, according to equation (1).

$$d_i = (w_{i,1}, w_{i,2}, \ldots, w_{i,j}, w_{i,t}) \tag{1}$$

For a document collection including $n$ documents and $t$ features (terms), VSM represents the text dataset as follows:

$$VSM = \begin{bmatrix} w_{1,1} & w_{1,2} & \cdots & w_{1,(t-1)} & w_{1,t} \\ \cdots & \cdots & \cdots & \cdots & \cdots \\ \vdots & \vdots & \ddots & \vdots & \vdots \\ \vdots & \vdots & \ddots & \vdots & \vdots \\ w_{(n-1),1} & w_{(n-1),2} & \cdots & \cdots & w_{(n-1),t} \\ w_{n,1} & w_{n,2} & \cdots & w_{n,(t-1)} & w_{n,t} \end{bmatrix} \tag{2}$$

Where $n$ is the number of documents, $t$ is the number of all features, $d_i$ is the document number $i$, and $w_{i,j}$ is the weight of the $j^{th}$ feature (term) in document number $i$.

Various weighting schemes are introduced in the literature; among these methods, Term Frequency-Inverse Document Frequency (TF-IDF) is widely employed to calculate term weights based on their frequency in documents. In this work, TF-IDF is used to represent documents in numeric form. This weighting scheme is mathematically formulated as equations (3) and (4).

$$w_{i,j} = tf(i,j) * idf(i,j) \tag{3}$$

$$idf(i,j) = \log(n/df(j)) \tag{4}$$

Where $w_{i,j}$ is the weight of the $j^{th}$ feature (term) in document number $i$, $tf(i,j)$ represents the frequency of term $j$ in document number $i$. Inverse document frequency ($idf(i,j)$), calculated according to equation (4), indicates the amount of information the word provides across all documents and determines whether the word is common or rare. $df(j)$ is the number of documents that contain the word $j$.

### 3.2. Feature selection using a hybrid algorithm of TLBO and GWO

Feature selection is an optimization problem looking for the most informative subset of features. As mentioned in previous sections, feature selection is applied document by document in the proposed method. Given the set of original features of document number $i$ as $F_i = f_{i,1}, f_{i,2}, \ldots, f_{i,j}, \ldots, f_{i,t}$, the solution of feature selection for document number $i$ is shown as $SF_i = sf_{i,1}, sf_{i,2}, \ldots, sf_{i,j}, \ldots, sf_{i,t}$. If $sf_{i,j} = 1$, the $j^{th}$ feature is selected as an informative feature, and $sf_{i,j} = 0$ means that the $j^{th}$ feature is not selected for document number $i$.

The details of the proposed hybrid algorithm are discussed in the following subsections.

### 3.2.1. Solution representation

The feature selection problem is formulated as an optimization problem, and a hybrid meta-heuristic algorithm is designed to find the most optimal solution (a subset of features). The optimization algorithm starts with several random solutions represented as binary vectors, as shown in Fig. 1. Each solution illustrates a candidate subset of features. In this representation, each solution contains $t$ (number of all features in the original dataset) positions in which each position indicates one feature. If the position number $j$ equals 0, the $j^{th}$ feature is not selected in this solution, and if this position equals 1, the $j^{th}$ feature is selected.

| 1 | 1 | 0 | 0 | 1 | 0 | 1 | 1 | 1 | 0 |
|---|---|---|---|---|---|---|---|---|---|

**Fig. 1** example of solution representation for feature selection

### 3.2.2. Teaching-Learning-Based Optimization (TLBO)

TLBO, proposed by Rao et al. in 2016, has been inspired by the teaching and learning process of the class [38]. This algorithm has two main stages, which are explained in the following subsections.

1) **Teacher phase**

In the teacher stage, the best solution is considered as a teacher, and the other solutions (students) are enhanced by approaching the teacher; a teacher can move the mean of the class to some extent. In iteration $i$, the teacher tries to move the mean ($M_i$) towards its level according to equation (5).

$$Difference\_mean_i = r_i(M_{new} - T_F M_i) \qquad (5)$$

Where $T_F$ represents the teaching factor and can be either 1 or 2, and $r_i$ is a random value in [0,1]. The other solutions (students) can learn based on the calculated $Difference\_mean_i$ and improve according to equation (6).

$$X_{new,i} = X_{old,i} + Difference\_mean_i \qquad (6)$$

After computing the fitness function, if the new solutions are better than the old ones, they will be replaced.

2) **Learner phase**

The next stage is the learner phase, in which for each solution $i$, another solution $j$ is chosen randomly. If $j$ is better than $i$, $i$ will interact with $j$ and learn from it. Otherwise, $i$ will move in the opposite direction. Hence the position of $i$ will be updated based on equation (7).

$$X_{new,i} = \begin{cases} X_{old,i} + r_i(X_i - X_j) & if\ f(X_i) < f(X_j) \\ X_{old,i} + r_i(X_j - X_i) & if\ f(X_j) < f(X_i) \end{cases} \quad (7)$$

It is worth noting that equation (7) is defined based on a minimization problem.

### 3.2.3. Grey Wolf Optimizer (GWO)

GWO is a novel meta-heuristic algorithm proposed by Mirjalili et al. in 2014 [29]. GWO is inspired by the grey wolves' leadership and hunting mechanism in which the three best solutions are selected as $\alpha, \beta$, and $\delta$ wolves. Afterwards, the selected leaders guide others ($\omega$ wolves) towards the prey. Three steps of hunting are defined and formulated as follows:

1) **Encircling**

The first action of grey wolves' haunting is to encircle the prey, which can be mathematically modeled according to equations (8) and (9).

$$\vec{D} = |\vec{C}.\vec{X}_P(t) - \vec{X}(t)| \quad (8)$$

$$\vec{X}(t+1) = \vec{X}_P(t) - \vec{A}.\vec{D} \quad (9)$$

Where $t$ is the number of the current iteration, $\vec{X}_P$ is the position of the prey, and $\vec{X}$ indicates the grey wolf's position. $\vec{C}$ and $\vec{A}$ are coefficient vectors and can be calculated based on equations (10) and (11).

$$\vec{A} = 2\vec{a}.\vec{r_1} - \vec{a} \quad (10)$$

$$\vec{C} = 2.\vec{r_2} \quad (11)$$

Where $\vec{r_1}, \vec{r_2}$ are random vectors with values in [0,1]. $\vec{a}$ linearly decreases from 2 to 0 during the iterations, according to equation (12).

$$\vec{a} = 2 - 2 \times t/MaxIter \quad (12)$$

2) **Hunting**

In this step, it is assumed that $\alpha, \beta$, and $\delta$ wolves have information about the prey's location and are able to lead other wolves. Hence, the three best solutions are considered as leader wolves, and the positions of others are updated according to the leaders' position. This process can be mathematically formulated as equations (13)-(15).

$$\vec{D_\alpha} = |\vec{C_1}.\vec{X}_\alpha - \vec{X}|, \vec{D_\beta} = |\vec{C_2}.\vec{X}_\beta - \vec{X}|, \vec{D_\delta} = |\vec{C_3}.\vec{X}_\delta - \vec{X}| \quad (13)$$

$$\vec{X_1} = \vec{X_\alpha} - \vec{A_1}.(\vec{D_\alpha}), \vec{X_2} = \vec{X_\beta} - \vec{A_2}.(\vec{D_\beta}), \vec{X_3} = \vec{X_\delta} - \vec{A_3}.(\vec{D_\delta}) \quad (14)$$

$$\vec{X}(t+1) = \frac{\vec{X_1} + \vec{X_2} + \vec{X_3}}{3} \tag{15}$$

Where $\vec{X}_\alpha$, $\vec{X}_\beta$, and $\vec{X}_\delta$ are the position of the best solutions, and $\vec{A}$ vectors are calculated according to equation (10).

### 3) Attacking

As the final step, wolves attack the prey when it stops moving. Attacking is mathematically simulated by decreasing the $\vec{a}$ value. As mentioned above, $\vec{a}$ decreases from 2 to 0 during the iterations and keeps a balance between exploration and exploitation.

### 3.2.4. Hybrid TLBO-GWO algorithm

The proposed algorithm can be considered as a modification of TLBO. In the main TLBO **Error! Reference source not found.**], there are 2 phases called "teacher" and "learner." In the proposed approach, three stages are designed, namely "Teaching," "Interactive learning," and "Self-learning." Each phase is explained in the following parts.

### 1) Teaching

In this stage, like the main TLBO, the best solution is regarded as a teacher and improves (teaches) other solutions (students). The distinction is that in TLBO-GWO, the students learn from the teacher using a crossover operator. The applied operator is a uniform crossover to enhance the exploration capability of TLBO-GWO to prevent possible entrapment in the local optimum. This operator works as equation (16), as shown in Fig. 2.

$$\vec{X}_i(t+1) = \vec{r}.\vec{X}_i(t) + (1-\vec{r}).\vec{X}_T(t) \tag{16}$$

Where $\vec{X}_i(t+1)$ is the position of the $i^{th}$ student in iteration $t+1$, $\vec{X}_i(t)$ is the position of $i^{th}$ student in iteration $t$, $\vec{r}$ is a random vector with values in {0,1}, and $\vec{X}_T(t)$ is the position of the teacher in iteration $t$.

| $\vec{X}_i(t)$ | 0 | 1 | 1 | 1 | 0 | 1 | 0 | 0 | 1 | 1 |
|---|---|---|---|---|---|---|---|---|---|---|
| $\vec{X}_T(t)$ | 1 | 0 | 0 | 1 | 0 | 1 | 1 | 1 | 0 | 1 |
| $\vec{r}$ | 0 | 0 | 1 | 0 | 1 | 1 | 0 | 1 | 0 | 1 |
| $\vec{X}_i(t+1)$ | 1 | 0 | 1 | 1 | 0 | 1 | 1 | 0 | 0 | 1 |

Fig. 2 sample of uniform crossover

### 2) Interactive learning

In this stage, interactive learning is conducted based on a mechanism inspired by GWO, in which learning will be done based on an interaction between three elite students and others. To put it in another way, students will learn from the three best students chosen as $\alpha$, $\beta$, and $\delta$ leaders. Then the students' positions will be updated according to equations (10)-(15) using the GWO process discussed earlier.

3) Self-learning

This stage is added to the main TLBO to enhance its exploration and exploitation capability using a mutation operator. A recently introduced rank-based mutation [40] is employed in TLBO-GWO. In this adaptive mutation strategy, after evaluating the students based on the fitness function, the rank of each student is assigned based on their fitness. In a population of $N$ students, rank $N$ is dedicated to the best student (based on fitness function), and one is assigned to the weakest student. The mutation probability is calculated in the next step according to equation (17).

$$p = p_{max} * \left(1 - \frac{r-1}{N-1}\right) \qquad (17)$$

Where $p_{max}$ indicates the maximum mutation probability, $r$ is the student's rank, and $N$ is the number of all students. This approach ensures that the mutation probability of the best student is 0, and the maximum probability is dedicated to the weakest student.

3.2.5. Fitness Function

Evaluating the students is an indispensable part of the proposed algorithm. For this purpose, Mean Absolute Difference (MAD) is used as a fitness function. MAD calculates the fitness of each solution according to equation (18) based on the assigned weights by TF-IDF.

$$MAD_{X_i} = \frac{1}{a_i} \sum_{j=1}^{t} |x_{i,j} - \bar{x}_i| \qquad (18)$$

Where $MAD_{X_i}$ is the fitness of student $i$, $a_i$ is the number of selected features in document $i$, $x_{i,j}$ is the weight of feature $j$ in document $i$, and $t$ represents the number of all unique features in the original dataset. $\bar{x}_i$ is the mean value of vector $i$ and is calculated according to equation (19).

$$\bar{x}_i = \left(\frac{1}{a_i}\right) \sum_{j=1}^{t} x_{i,j} \qquad (19)$$

3.2.6. Binarization

As explained in the solution representation section, FS is defined as a binary optimization problem. Therefore, the sigmoid function, represented in Fig. 3, is used to transform the results of FS into a binary format. The solutions are modified in each dimension according to equation (20) to be represented in a binary form.

$$X_{i,j} = \begin{cases} 1 & if\ rand < S_{i,j} \\ 0 & Else \end{cases} \quad (20)$$

$rand$ is a random value in the range of [0,1], and $S_{i,j}$ is calculated based on equation (21), in which $x_{i,j}$ is the value of feature $j$ in solution $i$.

$$S_{i.j} = \frac{1}{e^{-x_{i,j}}} \quad (21)$$

The pseudo-code of the proposed algorithm is displayed in Algorithm 1. Moreover, Table *1* describes the parameters used in Algorithm 1.

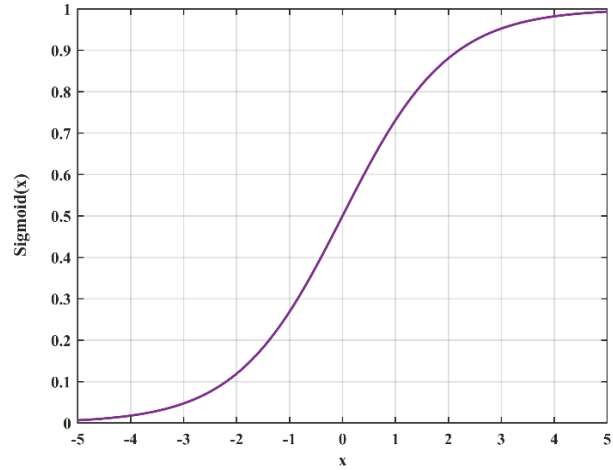

**Fig. 3** sigmoid function

---

**Algorithm 1.** Pseudo-code of TLBO-GWO for feature selection.

1: **Input:** Dataset, $N$, TLBO-GWO parameters ($Iter_{max}, N_{pop}, P_{max}$)
2: **Output:** Subset of features
3: **Algorithm**
4: for $i = 1$ to $N$ do
5:     Initialize $N_{pop}$ students randomly to produce $Pop$
7:     for $It = 1$ to $Iter_{max}$ do
8:         Evaluate students using equation (18)
9:         Select the best solution as $T$
10:         // Teaching
11:         for $j = 1$ to $N_{pop}$ do
12:             Crossover $Pop$ ($j$) and $T$ using equation (16)
13:             Evaluate the crossover's solution using equation (18)
14:             Update $Pop$ ($j$) if the crossover's solution is better
15:         end for
16:         // Interactive learning
17:         for $k = 1$ to $N_{pop}$ do
18:             Evaluate students using equation (18)

19:        Find the three best students as $\alpha, \beta, \delta$
20:        Update $Pop\ (k)$ by equations (10)-(15)
21:     **end for**
22:     Sort students based on fitness
23:     **// Self-learning**
24:     **for** $l = 1$ to $N_{pop}$ **do**
25:        Find the rank of $Pop\ (l)$ and calculate $p(l)$ based on equation (17)
26:        Mutate $Pop\ (l)$ with $p(l)$ probability
27:        Evaluate the mutation's solution using equation (18)
28:        Update $Pop\ (l)$ if the mutation's solution is better
29:     **end for**
30:   **end for**
31: **end for**
32: Return a new subset of features

Table 1  Description of parameters

| Parameters | Description |
| --- | --- |
| $N$ | Number of text documents in the dataset |
| $Iter_{max}$ | Maximum number of iterations |
| $N_{pop}$ | Number of population members (students) |
| $p_{max}$ | Maximum mutation probability |
| $Pop$ | Students' population |
| $T$ | Teacher |
| $Pop\ (j)$ | $j^{th}$ member of population |
| $p(l)$ | mutation probability of student $l$ |

### 3.3. Text clustering technique

The updated dataset, including the most informative features, should be clustered at this stage. According to the literature and experiments, K-means is found as an appropriate clustering technique. Given $D = d_1, d_2, \ldots, d_j, \ldots, d_n$ as a text dataset of $n$ documents and $d_i = w_{i,1} + w_{i,2} + w_{i,3} + \cdots + w_{i,t}$ as document number $i$; accordingly, each document is shown as a weighted vector of $t$ features. In this representation, $t$ points to the number of all features in $D$. The steps of this algorithm are presented as follows.

**Step 1:** Calculate each document's cosine similarity to others according to equation (22). Then calculate the mean similarity of each document to other documents. Consider the k most similar documents as cluster centroids.

$$cos(d_i, d_j) = \frac{d_i \cdot d_j}{||d_i|| * ||d_j||} = \frac{\sum_{k=1}^{t} w(t_k, d_i) \times w(t_k, d_j)}{\sqrt{\sum_{k=1}^{t} w(t_k, d_i)^2} \sqrt{\sum_{k=1}^{t} w(t_k, d_j)^2}} \quad (22)$$

**Step 2:** Assign each document to the nearest centroid based on the calculated similarities.
**Step 3:** update the clusters' centroids by averaging the assigned documents.
**Step 4:** Calculate each document's similarity to the new centroids and assign it to the most similar centroid.

**Step 5:** If the termination condition is reached, stop and return k coherent clusters of documents and if the termination condition is not met, return to step 3 and continue.

A flowchart of the whole procedure is displayed in Fig. 4.

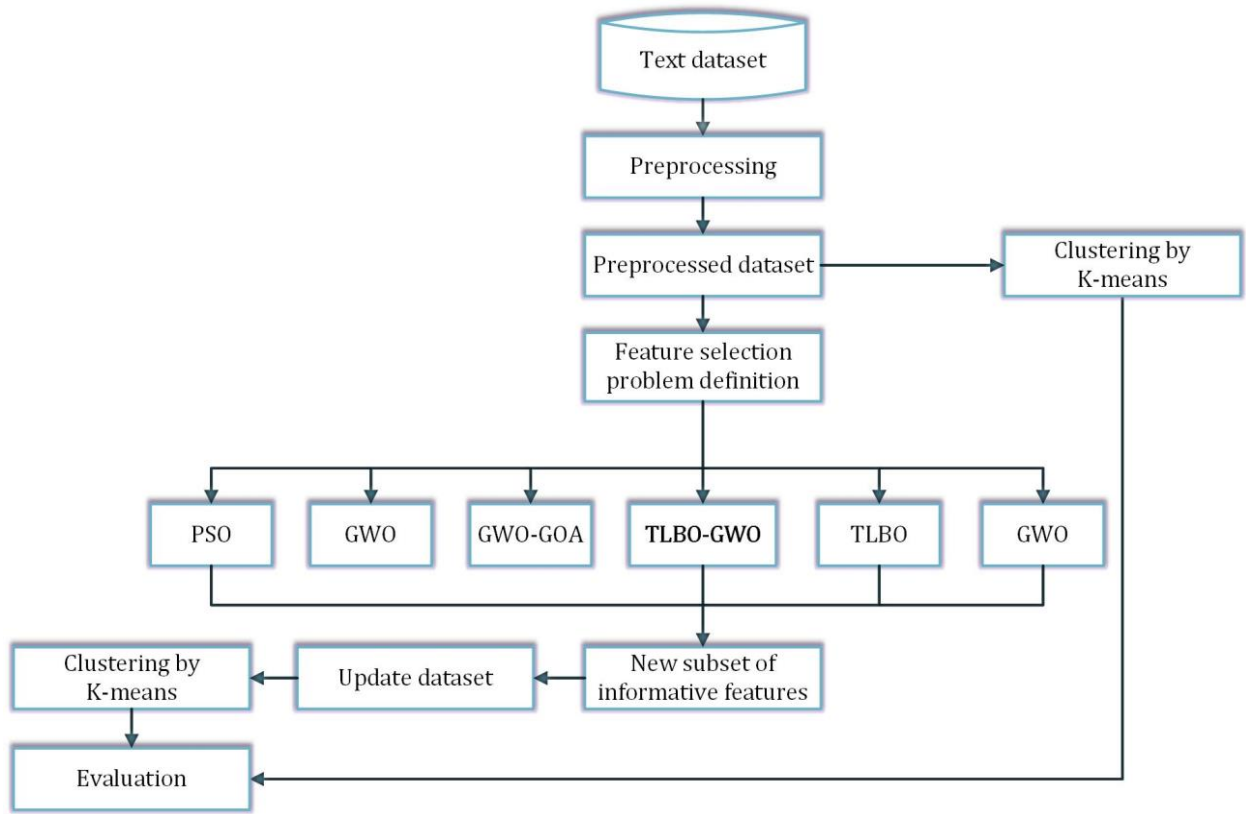

**Fig. 4** flowchart of the proposed methodology

## 4. Experimental setup

The following subsections explain the characteristics of datasets and employed measures for clustering evaluation.

### 4.1. Datasets

Benchmark datasets available at the Laboratory of Computational Intelligence (LABIC)[1] are used to evaluate the effectiveness of the proposed algorithm. Table 2 shows the detail of the employed datasets.

**Table 2** characteristics of datasets

| Dataset | Name of the original dataset | Number of documents | Number of terms | Number of clusters |
|---------|------------------------------|---------------------|-----------------|--------------------|
| DS1 | 20 Newsgroups | 100 | 1435 | 3 |
| DS2 | Reuters-21578 | 200 | 2038 | 3 |
| DS3 | Reuters-21578 | 200 | 2543 | 4 |
| DS4 | Reuters-21578 | 500 | 3518 | 6 |
| DS5 | DMOZ-Health | 1000 | 2299 | 10 |

---
[1] http://sites.labic.icmc.usp.br/text_collections/

| | | | | | |
|---|---|---|---|---|---|
| DS6 | 20 Newsgroups | 5000 | 13599 | 11 | |

### 4.2. Evaluation measures

Accuracy, precision, recall, and F-measure are employed as clustering evaluation measures. These measures are calculated according to equations (23)-(26).

$$Accuracy = \frac{TP + TN}{TP + TN + FP + FN} \tag{23}$$

$$Precision = \frac{TP}{TP + FP} \tag{24}$$

$$Recall = \frac{TP}{TP + FN} \tag{25}$$

$$F - Measure = \frac{2PR}{P + R} \tag{26}$$

### 4.3. Parameter setting

As mentioned earlier, three comparative algorithms [3], [6], [28] and the main TLBO [39] and GWO [29] are implemented to be compared with TLBO-GWO. The values of the common parameters are decided to be the same. Therefore, the maximum number of iterations and population number are set at 500 and 30, respectively. It is preferred to set the parameters of other algorithms according to their original papers. However, a parameter setting is carried out for algorithms with unclearly stated values. All parameters are described in Table 3.

Table 3  Parameter setting of the algorithms

| Algorithm | Parameters |
|---|---|
| PSO-GA | $C_1 = 2.2, C_2 = 2.2, Cr = 0.9, Mu = 0.2$ |
| PSO | $C_1=1.8, C_2=1.8$ |
| GWO-GOA | $C_{min} = 0.00001, C_{max} = 1, Cr = 0.75, Mu = 0.15$ |
| TLBO-GWO | $P_{max} = 0.08$ |
| TLBO | - |
| GWO | - |

## 5. Results and discussion

This section evaluates the results based on clustering measures, convergence behavior, and feature reduction ratio. Eventually, to ensure the effectiveness of the proposed algorithm in terms of clustering measures, statistical analysis is conducted.

### 5.1. Clustering measures

Table 4 compares the results of TLBO-GWO and five comparative algorithms in terms of the mentioned clustering evaluation measures. PSO-GA [6], PSO [28], GWO-GOA [3], TLBO [39], GWO [29], and TLBO-GWO are simulated and executed on six datasets to find an optimal subset of features. The results are calculated by averaging over twenty runs to compare the algorithms fairly. The iteration number of the K-means clustering algorithm is specified experimentally and set at 50. As shown in Table 4, in terms of precision for all datasets, TLBO-GWO outperformed the other algorithms. In addition, except for DS3, where GWO-GOA showed better performance, TLBO-GWO had the highest accuracy among other implemented algorithms. While for DS4 main TLBO contributed to the best results of recall and F-measure, in other datasets, the figure for TLBO-GWO was the highest. Furthermore, it is obvious that TLBO-GWO performed significantly better in comparison with K-means (without feature selection). Overall, the results indicate that the proposed algorithm surpasses other implemented algorithms in terms of clustering measures.

**Table 4** Evaluation of algorithms based on clustering measures

| Dataset | Measure | PSO-GA | PSO | TLBO | GWO | GWO-GOA | K-means | **TLBO-GWO** |
|---|---|---|---|---|---|---|---|---|
| DS1 | Accuracy | 0.2905 | 0.326 | 0.3075 | 0.2805 | 0.286 | 0.2675 | **0.393** |
| | Precision | 0.3028 | 0.3217 | 0.306 | 0.2888 | 0.2854 | 0.2705 | **0.3475** |
| | Recall | 0.2944 | 0.3198 | 0.303 | 0.2859 | 0.2538 | 0.2629 | **0.3536** |
| | F-Measure | 0.298 | 0.3203 | 0.3041 | 0.2862 | 0.263 | 0.2683 | **0.3465** |
| DS2 | Accuracy | 0.3545 | 0.3775 | 0.3725 | 0.36 | 0.3975 | 0.215 | **0.4932** |
| | Precision | 0.308 | 0.3359 | 0.3708 | 0.3821 | 0.3734 | 0.3199 | **0.3926** |
| | Recall | 0.3661 | 0.3618 | 0.3601 | 0.3632 | 0.3371 | 0.2584 | **0.3662** |
| | F-Measure | 0.3294 | 0.3466 | 0.355 | 0.3686 | 0.3474 | 0.2775 | **0.3744** |
| DS3 | Accuracy | 0.3046 | 0.2939 | 0.2708 | 0.3287 | **0.5235** | 0.2289 | 0.5054 |
| | Precision | 0.3044 | 0.2888 | 0.2776 | 0.2998 | 0.3085 | 0.2579 | **0.3176** |
| | Recall | 0.2809 | 0.2502 | 0.2685 | 0.2799 | 0.2988 | 0.2505 | **0.326** |
| | F-Measure | 0.2905 | 0.2756 | 0.2711 | 0.2884 | 0.302 | 0.2469 | **0.3174** |
| DS4 | Accuracy | 0.3304 | 0.2961 | 0.3821 | 0.3454 | 0.2883 | 0.243 | **0.442** |
| | Precision | 0.3325 | 0.3121 | 0.3548 | 0.3448 | 0.3442 | 0.2141 | **0.3978** |
| | Recall | 0.3407 | 0.3184 | **0.4038** | 0.3544 | 0.3058 | 0.2444 | 0.3503 |
| | F-Measure | 0.332 | 0.3046 | **0.3704** | 0.344 | 0.3171 | 0.2506 | 0.3688 |
| DS5 | Accuracy | 0.2643 | 0.2437 | 0.209 | 0.278 | 0.2549 | 0.1766 | **0.2749** |
| | Precision | 0.1891 | 0.2077 | 0.177 | 0.1929 | 0.1801 | 0.1637 | **0.2368** |
| | Recall | 0.2153 | 0.2054 | 0.185 | 0.2288 | 0.188 | 0.1543 | **0.2639** |
| | F-Measure | 0.1967 | 0.2009 | 0.1786 | 0.2038 | 0.1796 | 0.1553 | **0.2467** |
| DS6 | Accuracy | 0.2401 | 0.241 | 0.2393 | 0.2553 | 0.3428 | 0.2485 | **0.3466** |
| | Precision | 0.2227 | 0.2368 | 0.2238 | 0.2419 | 0.2403 | 0.2255 | **0.2592** |
| | Recall | 0.2341 | 0.2338 | 0.2315 | 0.2469 | 0.2431 | 0.2286 | **0.263** |
| | F-Measure | 0.2325 | 0.2347 | 0.2253 | 0.2439 | 0.2384 | 0.2257 | **0.2601** |

## 5.2. Convergence

The proposed algorithm is also evaluated by comparing its convergence behavior with other mentioned algorithms. The capability to maximize the fitness function (MAD) is investigated over 500 iterations, As shown in Fig. 5 to Fig. 10. It is clearly observed from the figures that in all six datasets, TLBO-GWO could approach optimal results with a better convergence than the other algorithms. In addition, TLBO-GWO could reach the highest level of fitness function. In almost all six datasets, PSO-GA [6], PSO [28], TLBO [39], and GWO [29] followed a similar trend, and their graphs were markedly close together.

In most cases, GWO-GOA peaked before iteration number 50 and remained constant during the next iterations. Therefore, although GWO-GOA was the most comparable algorithm to the proposed TLBO-GWO in terms of the final optimum result, it seems that GWO-GOA was trapped in the local optimum and suffered from premature convergence. For all datasets except DS5, GWO-GOA ranked second in maximization of the fitness function, whereas for DS5, this algorithm showed the weakest performance among all implemented algorithms. It is worth noting that the proposed algorithm showed a smoother convergent curve in all six datasets.

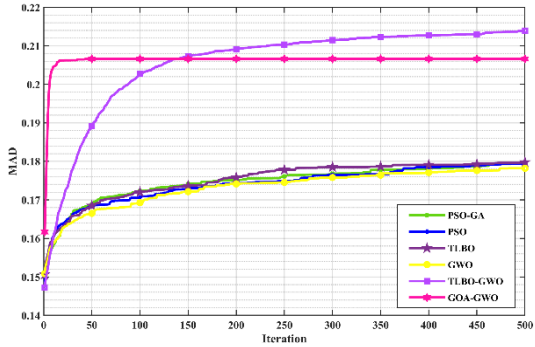

**Fig. 5** Convergence behavior for DS1

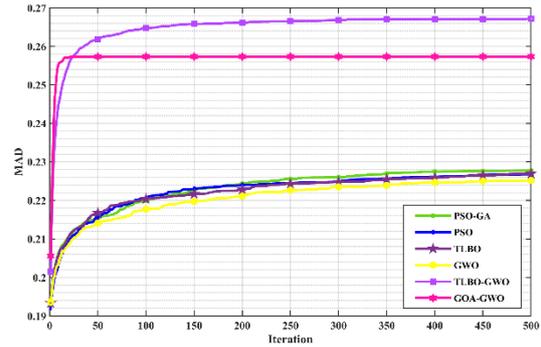

**Fig. 6** Convergence behavior for DS2

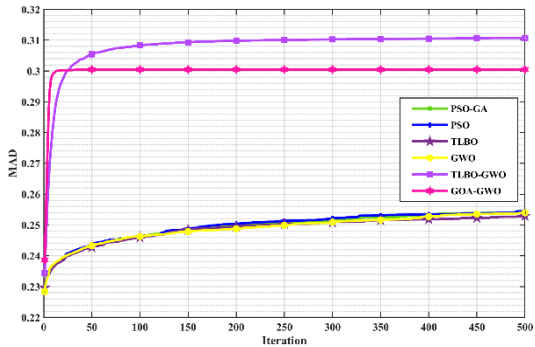

**Fig. 7** Convergence behavior for DS3

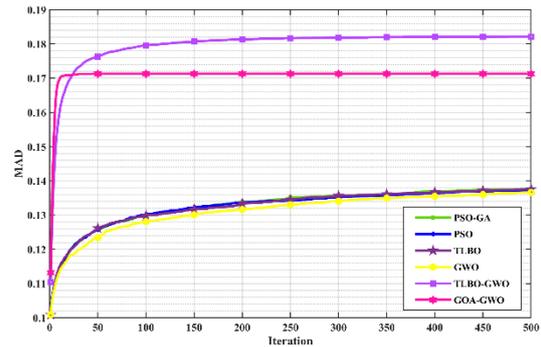

**Fig. 8** Convergence behavior for DS4

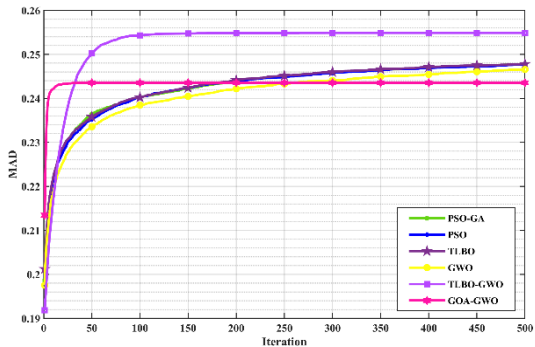

**Fig. 9** Convergence behavior for DS5

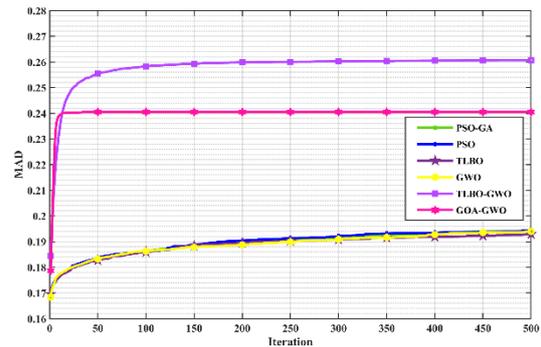

**Fig. 10** Convergence behavior for DS6

### 5.3. Feature reduction ratio

Another advantage of a FS algorithm is its ability to select the most informative and smallest subset of features. Table 5 displays the feature reduction ratios associated with each algorithm. Overall, according to the table, the proposed algorithm could reach the highest reduction ratio in all datasets. It also can be concluded that GWO-GOA was the most comparable algorithm to the proposed one in reducing feature space size. In addition, obtained data proves that PSO-GA [6], PSO [28], TLBO 39], and GWO [29] contributed to approximately the same reduction ratios, roughly 61% for DS1, 48% for DS2 and DS3, 43% for DS4, 55% for DS5 and nearly 46% for DS6. Furthermore, it can be inferred that GWO-GOA [3] and TLBO-GWO led to higher reduction ratios. For DS1, DS4 and DS5, these two algorithms showed the most similar feature reduction results at roughly 78%, 57%, and 62%, respectively. For DS2, DS3, and DS6, the reduction ratios of TLBO-GWO were the highest at 65%, 67%, and 64%, whereas the figures for GWO-GOA were lower at 62%, 63%, and %64, respectively.

Table 5 Feature reduction ratio

| Dataset | | original | PSO-GA | PSO | TLBO | GWO | GWO-GOA | **TLBO-GWO** |
|---|---|---|---|---|---|---|---|---|
| DS1 | Dimension | 1435 | 559 | 565 | 574 | 561 | 323 | 301 |
| | Reduction ratio | - | 61% | 61% | 60% | 61% | 77% | **79%** |
| DS2 | Dimension | 2038 | 1048 | 1052 | 1079 | 1054 | 769 | 721 |
| | Reduction ratio | - | 49% | 48% | 47% | 48% | 62% | **65%** |
| DS3 | Dimension | 2543 | 1366 | 1317 | 1331 | 1282 | 927 | 833 |
| | Reduction ratio | - | 46% | 48% | 48% | 50% | 63% | **67%** |
| DS4 | Dimension | 3518 | 2018 | 2007 | 2046 | 1983 | 1537 | 1502 |
| | Reduction ratio | - | 43% | 43% | 42% | 44% | 56% | **57%** |
| DS5 | Dimension | 2299 | 1027 | 1032 | 1035 | 1013 | 905 | 857 |
| | Reduction ratio | - | 55% | 55% | 55% | 56% | 61% | **63%** |
| DS6 | Dimension | 13599 | 7443 | 7508 | 7239 | 7058 | 5295 | 4927 |
| | Reduction ratio | - | 45% | 45% | 47% | 48% | 61% | **64%** |

### 5.4. Statistical analysis

In this section, to further investigate the effectiveness and superiority of the proposed algorithm, a comparison is conducted in terms of accuracy, precision, recall, and F-measure using statistical tests. Since a normal distribution was not observed in a few cases, the T-test might not have sufficient validity. Therefore, the Mann-Whitney test is also implemented as a non-parametric test to ensure the results. Tests are carried out on six datasets through 20 runs, and the confidence level threshold is set at 95%. The null hypothesis is that the performance of TLBO-GWO is similar or inferior to another algorithm (e.g., GWO-GOA). The alternative hypothesis is that TLBO-GWO is significantly better than another algorithm.

Table 6 represents the results of the tests, which provide *p-values*. If a *p-value* associated with a measure is less than 0.05, the null hypothesis is rejected, and the superiority of the proposed algorithm is concluded. Overall, it can be deduced from the results that TLBO-GWO performed considerably better in most cases. Besides, it can be inferred that T-test and Mann-Whitney test mostly achieved similar results. To illustrate, given the first dataset (DS1), both tests prove the proposed algorithm's accuracy superiority. Regarding precision, while the T-test proved that TLBO-GWO is better than all other algorithms, the Mann-Whitney test confirmed the superiority of TLBO-GWO over all algorithms except PSO, which had an equal or better performance. Both statistical tests demonstrated that TLBO-GWO had better performance than PSO-GA, PSO, TLBO, and GWO concerning recall and F-measure, whereas PSO had an equal or better performance than TLBO-GWO.

Table 6 Statistical results of T-test and Mann-Whitney test with α<0.05

| Dataset | Measure | T-test | | | | | | Mann-Whitney test | | | | | |
|---|---|---|---|---|---|---|---|---|---|---|---|---|---|
| | | PSO-GA | PSO | TLBO | GWO | GWO-GOA | k-means | PSO-GA | PSO | TLBO | GWO | GWO-GOA | k-means |
| DS1 | A | **0.000** | **0.000** | **0.000** | **0.000** | **0.000** | **0.000** | **0.000** | **0.000** | **0.000** | **0.000** | **0.000** | **0.000** |
| | P | **0.002** | **0.033** | **0.001** | **0.002** | **0.000** | **0.000** | **0.009** | 0.08 | **0.001** | **0.006** | **0.000** | **0.000** |
| | R | **0.008** | 0.081 | **0.013** | **0.005** | **0.000** | **0.000** | **0.015** | 0.090 | **0.021** | **0.01** | **0.000** | **0.001** |
| | F | **0.004** | 0.067 | **0.004** | **0.002** | **0.000** | **0.000** | **0.007** | 0.095 | **0.005** | **0.004** | **0.000** | **0.000** |
| DS2 | A | 0.058 | **0.011** | **0.008** | **0.003** | **0.025** | **0.000** | **0.008** | **0.022** | **0.005** | **0.002** | **0.008** | **0.000** |
| | P | **0.002** | **0.016** | 0.244 | 0.352 | 0.262 | **0.004** | **0.027** | **0.018** | 0.378 | 0.224 | 0.262 | **0.003** |
| | R | 0.499 | 0.437 | 0.414 | 0.458 | 0.169 | **0.000** | 0.622 | 0.357 | 0.662 | 0.473 | 0.441 | **0.001** |
| | F | **0.036** | 0.123 | 0.23 | 0.41 | 0.17 | **0.000** | 0.175 | 0.095 | 0.414 | 0.262 | 0.347 | **0.000** |
| DS3 | A | **0.000** | **0.000** | **0.000** | **0.000** | 0.699 | **0.000** | **0.000** | **0.000** | **0.000** | **0.000** | 0.692 | **0.000** |
| | P | 0.293 | 0.143 | **0.039** | 0.187 | 0.308 | **0.048** | 0.308 | 0.28 | **0.016** | 0.186 | 0.254 | **0.023** |
| | R | **0.023** | **0.007** | **0.005** | **0.017** | 0.114 | **0.004** | **0.043** | **0.005** | **0.003** | **0.023** | 0.22 | **0.004** |
| | F | 0.085 | **0.023** | **0.006** | **0.044** | 0.177 | **0.008** | 0.097 | **0.034** | **0.005** | 0.054 | 0.29 | **0.005** |
| DS4 | A | **0.004** | **0.001** | 0.103 | **0.013** | **0.001** | **0.000** | **0.001** | **0.001** | 0.07 | **0.009** | **0.001** | **0.001** |
| | P | **0.044** | **0.003** | 0.127 | 0.066 | **0.039** | **0.000** | **0.043** | **0.01** | 0.063 | **0.025** | **0.038** | **0.000** |
| | R | 0.398 | 0.212 | 0.911 | 0.542 | 0.106 | **0.008** | 0.398 | 0.308 | 0.91 | 0.72 | 0.245 | **0.006** |
| | F | 0.145 | **0.026** | 0.518 | 0.232 | **0.046** | **0.004** | 0.29 | 0.065 | 0.706 | 0.28 | 0.082 | **0.002** |
| DS5 | A | 0.4 | 0.216 | **0.037** | 0.533 | 0.319 | **0.012** | 0.436 | 0.276 | **0.028** | 0.533 | 0.328 | **0.001** |
| | P | **0.009** | 0.071 | **0.002** | **0.016** | **0.002** | **0.001** | **0.009** | **0.041** | **0.002** | **0.019** | **0.001** | **0.001** |
| | R | **0.013** | **0.008** | **0.001** | 0.053 | **0.001** | **0.000** | **0.012** | **0.007** | **0.001** | 0.074 | **0.002** | **0.000** |
| | F | **0.005** | **0.012** | **0.000** | **0.011** | **0.001** | **0.000** | **0.011** | **0.007** | **0.000** | **0.016** | **0.000** | **0.000** |
| DS6 | A | **0.000** | **0.000** | **0.000** | **0.000** | 0.419 | **0.000** | **0.000** | **0.000** | **0.000** | **0.000** | 0.559 | **0.000** |
| | P | **0.000** | **0.035** | **0.003** | **0.037** | **0.005** | **0.044** | **0.000** | **0.027** | **0.005** | **0.015** | **0.009** | **0.000** |
| | R | **0.017** | **0.018** | **0.021** | 0.102 | 0.127 | **0.035** | **0.038** | **0.038** | **0.034** | 0.162 | 0.07 | **0.006** |
| | F | **0.01** | **0.019** | **0.005** | 0.054 | **0.021** | **0.033** | **0.01** | **0.023** | **0.003** | **0.047** | 0.057 | **0.002** |

## 6. Conclusion

Due to the significant role of text clustering in organizing large text datasets, numerous works have been carried out to improve its performance and results. In this regard, removing noninformative and redundant features, known as feature selection, is one of the effective solutions. This paper proposed a modified version of Teaching-Learning-Based Optimization (TLBO) and designed a filter-based feature selection algorithm. TLBO was enhanced using Grey Wolf Optimizer (GWO) and Genetic Algorithm (GA) to prevent premature convergence and entrapment in a local optimum. Six benchmark datasets of different sizes were selected, and the Mean Absolute Difference (MAD) was employed as a fitness function. Three recently proposed comparative algorithms, simple TLBO, and main GWO were also implemented to be compared with the proposed algorithm. After selecting an informative subset of features, the K-means clustering algorithm was applied to cluster the updated text dataset. The results showed that TLBO-GWO could outperform the other algorithms in clustering evaluation measures, convergence behavior, and feature reduction ratio. The validity of the results was further investigated using the T-test and the Man-Whitney test. These statistical tests confirmed the superiority of TLBO-GWO in most cases.

Other novel meta-heuristics and their different hybrids can be examined for future works to solve the same problem. In addition, other fitness functions and weighting methods may lead to better results. In this work, six parts of benchmark datasets were used; different datasets can be tested in future works. Besides, other clustering algorithms can be tested based on a similar feature selection approach. Last but not least, the proposed modification of TLBO is worth implementing on different optimization problems and is likely to reach satisfactory results.

## Declarations

### Ethical Approval

Not applicable

### Competing interests

The authors have no relevant financial or non-financial interests to disclose.

### Authors' contributions

All authors contributed to the study's conception and design. Data collection, coding, and preparation of figures were performed by Mahsa Azarshab. Evaluations and statistical analysis were conducted by Mahsa Azarshab, Mohammad Fathian, and Babak Amiri. The first draft of the manuscript was written by Mahsa Azarshab and all authors commented on previous versions of the manuscript. All authors read and approved the final manuscript.

### Funding

This research did not receive any specific grant from funding agencies in the public, commercial, or not-for-profit sectors.

### Availability of data and materials

The datasets generated during and/or analyzed during the current study are available from the corresponding author upon reasonable request.

References

1. Fouchal S, Ahat M, Ben Amor S, et al (2013) Competitive clustering algorithms based on ultrametric properties. Journal of Computational Science 4:219–231. https://doi.org/10.1016/j.jocs.2011.11.004
2. Lu Y, Liang M, Ye Z, Cao L (2015) Improved particle swarm optimization algorithm and its application in text feature selection. Applied Soft Computing 35:629–636. https://doi.org/10.1016/j.asoc.2015.07.005
3. Purushothaman R, Rajagopalan SP, Dhandapani G (2020) Hybridizing Gray Wolf Optimization (GWO) with Grasshopper Optimization Algorithm (GOA) for text feature selection and clustering. Applied Soft Computing 96:106651. https://doi.org/10.1016/j.asoc.2020.106651
4. Vergara JR, Estévez PA (2013) A review of feature selection methods based on mutual information. Neural Computing and Applications 24:175–186. https://doi.org/10.1007/s00521-013-1368-0
5. Uğuz H (2011) A two-stage feature selection method for text categorization by using information gain, principal component analysis and genetic algorithm. Knowledge-Based Systems 24:1024–1032. https://doi.org/10.1016/j.knosys.2011.04.014
6. Bharti KK, Singh PK (2015) Hybrid dimension reduction by integrating feature selection with feature extraction method for text clustering. Expert Systems with Applications 42:3105–3114. https://doi.org/10.1016/j.eswa.2014.11.038
7. Abualigah LM, Khader AT (2017) Unsupervised text feature selection technique based on hybrid particle swarm optimization algorithm with genetic operators for the text clustering. The Journal of Supercomputing 73:4773–4795. https://doi.org/10.1007/s11227-017-2046-2
8. Shang W, Huang H, Zhu H, et al (2007) A novel feature selection algorithm for text categorization. Expert Systems with Applications 33:1–5. https://doi.org/10.1016/j.eswa.2006.04.001
9. Mafarja M, Jarrar R, Ahmad S, Abusnaina AA (2018) Feature selection using binary particle swarm optimization with time varying inertia weight strategies. Proceedings of the 2nd International Conference on Future Networks and Distributed Systems. https://doi.org/10.1145/3231053.3231071
10. Mafarja M, Sabar NR (2018) Rank based binary particle swarm optimisation for feature selection in classification. Proceedings of the 2nd International Conference on Future Networks and Distributed Systems. https://doi.org/10.1145/3231053.3231072
11. Allam M, Nandhini M (2022) Optimal feature selection using binary teaching learning based optimization algorithm. Journal of King Saud University - Computer and Information Sciences 34:329–341. https://doi.org/10.1016/j.jksuci.2018.12.001
12. Mirjalili S, Lewis A (2016) The whale optimization algorithm. Advances in Engineering Software 95:51–67. https://doi.org/10.1016/j.advengsoft.2016.01.008

13. Mafarja M, Mirjalili S (2018) Whale optimization approaches for wrapper feature selection. Applied Soft Computing 62:441–453. https://doi.org/10.1016/j.asoc.2017.11.006
14. Alam WU, Ashraf A (2019) Improved Binary Bat Algorithm for Feature Selection. Thesis
15. Rashedi E, Nezamabadi-pour H, Saryazdi S (2009) GSA: A gravitational search algorithm. Information Sciences 179:2232–2248. https://doi.org/10.1016/j.ins.2009.03.004
16. Prakash J, Singh PK (2017) Gravitational search algorithm and K-means for simultaneous feature selection and data clustering: A multi-objective approach. Soft Computing 23:2083–2100. https://doi.org/10.1007/s00500-017-2923-x
17. Mirjalili S (2016) SCA: A sine cosine algorithm for solving optimization problems. Knowledge-Based Systems 96:120–133. https://doi.org/10.1016/j.knosys.2015.12.022
18. Abualigah L, Dulaimi AJ (2021) A novel feature selection method for data mining tasks using hybrid sine cosine algorithm and genetic algorithm. Cluster Computing 24:2161–2176. https://doi.org/10.1007/s10586-021-03254-y
19. Nouri-Moghaddam B, Ghazanfari M, Fathian M (2021) A novel multi-objective forest optimization algorithm for Wrapper Feature Selection. Expert Systems with Applications 175:114737. https://doi.org/10.1016/j.eswa.2021.114737
20. Ghaemi M, Feizi-Derakhshi M-R (2014) Forest optimization algorithm. Expert Systems with Applications 41:6676–6687. https://doi.org/10.1016/j.eswa.2014.05.009
21. Samieiyan B, MohammadiNasab P, Mollaei MA, et al (2022) Novel optimized crow search algorithm for feature selection. Expert Systems with Applications 204:117486. https://doi.org/10.1016/j.eswa.2022.117486
22. Askarzadeh A (2016) A novel metaheuristic method for solving constrained engineering optimization problems: Crow search algorithm. Computers & Structures 169:1–12. https://doi.org/10.1016/j.compstruc.2016.03.001
23. Khosravi H, Amiri B, Yazdanjue N, Babaiyan V (2022) An improved group teaching optimization algorithm based on local search and chaotic map for feature selection in high-dimensional data. Expert Systems with Applications 204:117493. https://doi.org/10.1016/j.eswa.2022.117493
24. Zhang Y, Jin Z (2020) Group Teaching Optimization Algorithm: A novel metaheuristic method for solving global optimization problems. Expert Systems with Applications 148:113246. https://doi.org/10.1016/j.eswa.2020.113246
25. Bharti KK, Singh PK (2014) A three-stage unsupervised dimension reduction method for text clustering. Journal of Computational Science 5:156–169. https://doi.org/10.1016/j.jocs.2013.11.007
26. Shamsinejadbabki P, Saraee M (2011) A new unsupervised feature selection method for text clustering based on genetic algorithms. Journal of Intelligent Information Systems 38:669–684. https://doi.org/10.1007/s10844-011-0172-5
27. Hong S-S, Lee W, Han M-M (2015) The Feature Selection Method based on Genetic Algorithm for Efficient of Text Clustering and Text Classification . International Journal of Advances in Soft Computing and its Applications

28. Abualigah LM, Khader AT, Hanandeh ES (2018) A new feature selection method to improve the document clustering using particle swarm optimization algorithm. Journal of Computational Science 25:456–466. https://doi.org/10.1016/j.jocs.2017.07.018
29. Mirjalili S, Mirjalili SM, Lewis A (2014) Grey Wolf optimizer. Advances in Engineering Software 69:46–61. https://doi.org/10.1016/j.eswa.2016.06.004
30. Saremi S, Mirjalili S, Lewis A (2017) Grasshopper optimisation algorithm: Theory and application. Advances in Engineering Software 105:30–47. https://doi.org/10.1016/j.advengsoft.2017.01.004
31. Gandomi AH, Alavi AH (2012) Krill herd: A new bio-inspired optimization algorithm. Communications in Nonlinear Science and Numerical Simulation 17:4831–4845. https://doi.org/10.1016/j.cnsns.2012.05.010
32. Abualigah L, Alsalibi B, Shehab M, et al (2020) A parallel hybrid krill herd algorithm for feature selection. International Journal of Machine Learning and Cybernetics 12:783–806. https://doi.org/10.1007/s13042-020-01202-7
33. Kushwaha N, Pant M (2018) Link based BPSO for feature selection in Big Data Text Clustering. Future Generation Computer Systems 82:190–199. https://doi.org/10.1016/j.future.2017.12.005
34. Zong Woo Geem, Joong Hoon Kim, Loganathan GV (2001) A new heuristic optimization algorithm: Harmony search. SIMULATION 76:60–68. https://doi.org/10.1177/003754970107600201
35. Abualigah LM, Khader AT, Al-Betar MA (2016) Unsupervised feature selection technique based on harmony search algorithm for improving the text clustering. 2016 7th International Conference on Computer Science and Information Technology (CSIT). https://doi.org/10.1109/CSIT.2016.7549456
36. Wu Y, Gao X-Z, Zenger K (2011) Knowledge-based artificial fish-swarm algorithm. IFAC Proceedings Volumes 44:14705–14710. https://doi.org/10.3182/20110828-6-IT-1002.02813
37. Thiyagarajan D, Shanthi N (2017) A modified multi objective heuristic for effective feature selection in text classification. Cluster Computing 22:10625–10635. https://doi.org/10.1007/s10586-017-1150-7
38. Rao RV, Savsani VJ, Vakharia DP (2011) Teaching–learning-based optimization: A novel method for constrained mechanical design optimization problems. Computer-Aided Design 43:303–315. https://doi.org/10.1016/j.cad.2010.12.015
39. Thawkar S (2021) A hybrid model using teaching–learning-based optimization and SALP SWARM algorithm for feature selection and classification in Digital Mammography. Journal of Ambient Intelligence and Humanized Computing 12:8793–8808. https://doi.org/10.1007/s12652-020-02662-z
40. Rao RV (2015) Teaching-learning-based optimization algorithm. Teaching Learning Based Optimization Algorithm 9–39. https://doi.org/10.1007/978-3-319-22732-0_2

41. Basak A (2020) A rank based adaptive mutation in genetic algorithm. International Journal of Computer Applications 175:49–55. https://doi.org/10.5120/ijca2020920572